\documentclass[runningheads]{llncs}
\usepackage[T1]{fontenc}
\usepackage[utf8]{inputenc}
\usepackage{amsmath,amssymb}
\usepackage{graphicx}
\usepackage{url}
\usepackage{multirow}
\usepackage{booktabs}
\usepackage{siunitx}
\usepackage{xcolor}
\usepackage{placeins}
\begin{document}
\title{Boosting Ultrasound Image Classification via Attribute-Guided Dual-Branch Framework}
\titlerunning{Boosting Ultrasound Image Classification via AttrGuide}
\author{Bo Zhao\inst{1}\textsuperscript{*}, Yapeng Li\inst{1}\textsuperscript{*}, Juhua Liu\inst{1}\textsuperscript{(\(\boxtimes\))}, and
Bo Du\inst{1}}
\authorrunning{B. Zhao et al.}
\institute{School of Computer Science, National Engineering Research Center for Multimedia
Software, Institute of Artificial Intelligence, Hubei Key Laboratory of Multimedia and
Network Communication Engineering, Wuhan University, Wuhan, China\\
\email{liujuhua@whu.edu.cn}}
\maketitle
\begingroup
\renewcommand{\thefootnote}{}
\footnotetext{* Equal contribution. \quad \(\boxtimes\) Corresponding author.}
\endgroup
\begin{abstract}
Ultrasound image classification is essential for computer-aided diagnosis. However, current methods often neglect clinical priors, leading to poor generalization in challenging scenarios and a lack of interpretability that limits clinical adoption. To address these issues, we aim to develop a medical-prior module that can be seamlessly integrated into existing pipelines to enhance both diagnostic performance and interpretability. In this paper, we propose an attribute-guided dual-branch framework for ultrasound classification that introduces domain-agnostic medical attribute priors, improving generalization while offering interpretable evidence. Specifically, a baseline branch follows conventional architectures and predicts image categories via a fully connected classifier. An attribute-guided branch injects domain-agnostic attributes as priors and produces human-interpretable decision cues. Finally, an adaptive decision module fuses the two branches in a data-dependent manner to yield the final prediction. Experiments across diverse ultrasound classification tasks demonstrate that our approach can be integrated into multiple backbones and state-of-the-art methods with low overhead, consistently improving accuracy and interpretability.
Code is available at: \url{https://github.com/zhaobo253-crypto/AttrGuide}.
\end{abstract}

\section{Introduction}
Ultrasound imaging is widely used in routine clinical practice because it is non-invasive, real-time, portable, and cost-effective\cite{ref1,ref2}. It supports screening, follow-up, and point-of-care assessment across fetal, breast, and other organ systems\cite{ref1,ref2,ref14,ref15,ref30}. However, ultrasound images often suffer from speckle noise, low contrast, operator dependence, and large appearance variations across devices and acquisition settings, motivating extensive deep learning-based analysis\cite{ref4,ref5}. These factors make reliable interpretation difficult even for experienced clinicians and raise concerns about model generalization under distribution shift\cite{ref10}. Therefore, accurate ultrasound image classification is important for computer-aided diagnosis, including fetal standard-plane recognition and breast lesion assessment\cite{ref1,ref2,ref3}.

\begin{figure}[t]
\centering
\includegraphics[width=0.95\textwidth]{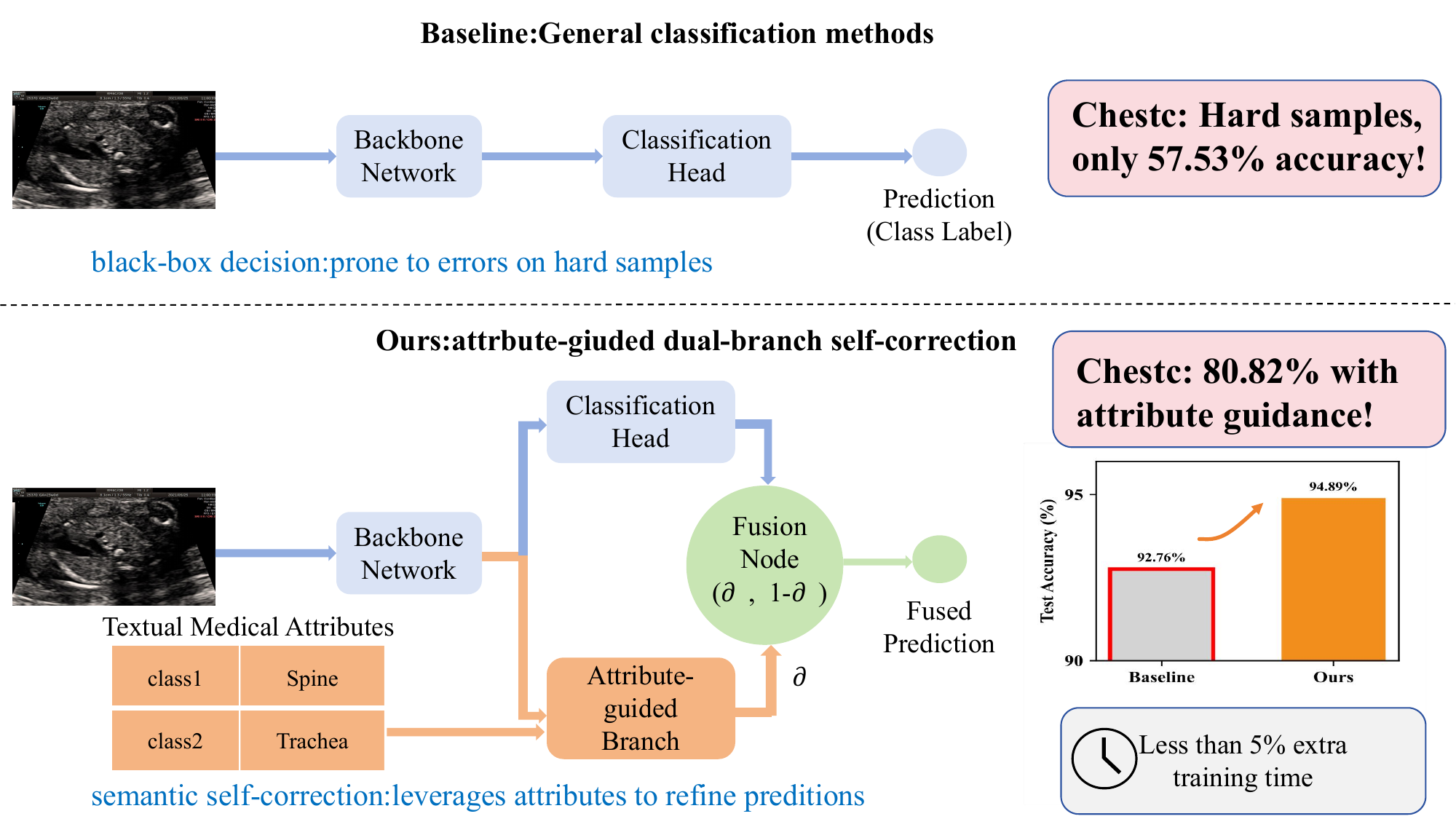}
\caption{Motivation. Traditional classifiers fail on hard samples; AttrGuide adds an attribute-guided branch for semantic self-correction with negligible overhead.}
\label{fig:motivation}
\end{figure}

Existing ultrasound classification methods mainly follow two paradigms. Conventional representation learning methods use transfer learning or self-supervised pretraining to extract discriminative features\cite{ref6,ref10,ref11,ref12,ref13,ref14,ref15,ref30}. Although effective on benchmark datasets, these ``black-box'' models may overfit superficial textures rather than clinically meaningful attributes such as echo patterns, boundary cues, and internal structures. This can lead to misclassifications on hard cases (Fig.~\ref{fig:motivation}) and limits clinical trust because the prediction is difficult to verify. Interpretable and attribute-guided models, including Concept Bottleneck Models and Vision-Language Models, introduce semantic concepts to align model reasoning with clinical cognition\cite{ref16,ref17,ref18,ref20,ref21,ref22,ref23,ref24,ref32}. Yet their dependence on dense annotations, task-specific concept labels, or complex prompt-tuning\cite{ref7,ref8,ref9} limits lightweight adaptation and increases computational cost. Thus, improving both performance and interpretability without heavy annotation burdens remains challenging.

Motivated by this, we aim to develop a plug-and-play module for existing architectures\cite{ref3} that improves diagnostic accuracy and transparency while preserving their original training pipelines. Two issues are central: how to transform domain-agnostic medical priors into a structured representation that can guide feature learning, and how to adaptively combine global semantic predictions with attribute-based cues across diverse clinical cases. A practical solution should provide interpretable evidence without requiring dense attribute annotation for every image.

In this paper, we propose AttrGuide, an attribute-guided dual-branch framework that can be integrated into existing classifiers with negligible overhead. The original architecture serves as the baseline branch, while a parallel attribute-guided branch injects clinical priors to bridge visual features and medical knowledge. It constructs a lightweight semantic pathway from predefined medical attributes and reuses the backbone features for attribute matching. An adaptive fusion module then combines baseline logits with prior-informed attribute cues in a data-dependent manner, enabling low-cost self-correction and improving both generalization and interpretability. Since the branch reuses backbone features and performs decision-level fusion, the framework requires only minimal structural modification.

To summarize, the main contributions of this paper are:
\begin{itemize}
\item \textbf{Plug-and-play framework:} We propose an attribute-guided dual-branch framework that can be seamlessly integrated into existing models to enhance both robustness and interpretability.
\item \textbf{Clinical-prior injection:} We design a dedicated attribute-guided branch that injects domain-agnostic medical priors to provide human-interpretable diagnostic evidence.
\item \textbf{Adaptive decision fusion:} We introduce an adaptive fusion module that dynamically reconciles global and attribute-based decisions to yield superior classification performance.
\end{itemize}
Extensive experiments show that AttrGuide can be integrated into various SOTA backbones, consistently improving performance and interpretability with negligible computational overhead.

\section{Method}

\begin{figure}[t]
\centering
\includegraphics[width=0.9\textwidth]{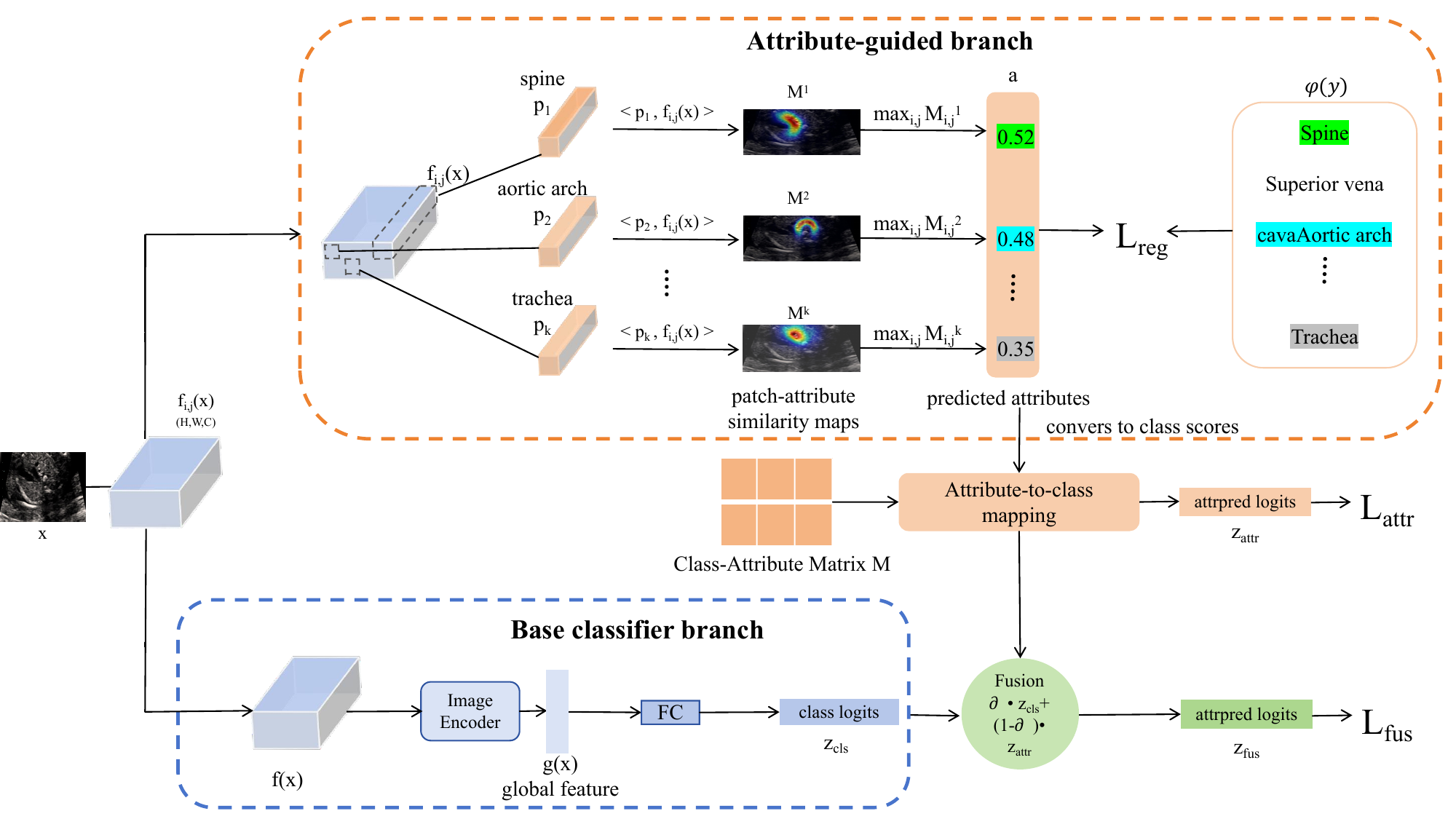}
\caption{Overview of AttrGuide. Our plug-in module can be seamlessly integrated into existing ultrasound classifiers by reusing their encoder feature maps: a baseline branch follows the original encoder--classifier design, while an attribute-guided branch matches the same features with CLIP-derived attribute prototypes, aggregates them into attribute scores, maps them to class scores via a fixed class--attribute matrix, and is jointly optimized with the baseline prediction through classification and attribute-based regularization losses.}
\label{fig:framework}
\end{figure}

We consider a labeled ultrasound dataset \(\mathcal{D}=\{(x_i,y_i)\}_{i=1}^N\), where \(x_i\) is an image and \(y_i \in \{1,\dots,C\}\) is the class label. We collect English attribute words \(\mathcal{A}=\{a_k\}_{k=1}^{K}\) and build a fixed class--attribute matrix \(M\in\{0,1\}^{C\times K}\), where \(M_{c,k}=1\) if class \(c\) is expected to exhibit attribute \(k\). Our goal is to learn a mapping \(f_\theta: x \mapsto z_{\text{fus}} \in \mathbb{R}^C\), where \(\theta\) denotes all learnable parameters and \(z_{\text{fus}}\) are fused logits that integrate a conventional baseline branch with an attribute-guided branch before softmax. The overall architecture (Figure~\ref{fig:framework}) treats any existing encoder+FC classifier as the baseline branch producing logits \(z_{\text{cls}}\), reuses its encoder features to drive an additional attribute-guided branch that produces attribute-based class logits \(z_{\text{attr}}\) via \(M\), and finally combines \(z_{\text{cls}}\) and \(z_{\text{attr}}\) through a lightweight fusion block at the decision level.

\subsection{Building the Medical Attribute Semantic Space}

We first construct an attribute semantic space on the text side. Let the pre-trained CLIP text encoder be \(T(\cdot)\)\cite{ref19}. For each attribute word \(a_k\), we build an ultrasound-specific prompt and encode it as \(e_k = T(\texttt{prompt}(a_k)) \in \mathbb{R}^{d}\), stacking all attribute embeddings into \(E = [e_1;\dots;e_K] \in \mathbb{R}^{K\times d}\).

The matrix \(E\) is precomputed and frozen. On the image side, the backbone \(B(\cdot)\) outputs local features \(v_i\) (ViT tokens or ResNet patches), which together with \(\{e_k\}\) are projected into a shared space and \(\ell_2\)-normalized as \(\tilde v_i=\mathrm{norm}(W_v v_i)\) and \(\tilde e_k=\mathrm{norm}(W_a e_k)\).

In this common space, we measure the correlation between each patch and each attribute via cosine similarity and aggregate patches to obtain per-attribute logits:
\begin{equation}
s_{k,i} = \cos(\tilde v_i,\tilde e_k),\quad
w_{k,i} = \mathrm{softmax}_i(\gamma\cdot s_{k,i}),\quad
\ell_k = \sum_{i=1}^{N} w_{k,i}\, s_{k,i}.
\label{eq:attr-logits}
\end{equation}
where \(\gamma\) is a scaling factor and \(\ell=[\ell_1,\dots,\ell_K]\) is the attribute prediction vector. Given the class--attribute matrix \(M\), we derive for each class \(y\) a binary target vector \(m_y\in\{0,1\}^{K}\) that encodes which attributes are expected to be present, and supervise attribute prediction with:
\begin{equation}
\mathcal{L}_{\text{attr-pred}}=\mathrm{BCEWithLogits}(\ell, m_{y}).
\label{eq:attr-loss}
\end{equation}
Beyond this attribute-prediction loss, we introduce a regularization term that further pulls the predicted attribute activations towards the same target vector \(m_y\) using both an element-wise penalty and a directional constraint in the semantic space:
\begin{equation}
\mathcal{L}_{\text{reg}} = \mathrm{MSE}(\sigma(\ell), m_y) + \bigl(1-\cos(\sigma(\ell), m_y)\bigr).
\label{eq:reg-loss}
\end{equation}

\subsection{Dual-Branch Adaptive Fusion Module}

\textbf{Baseline classification branch.} The global feature \(g\) from the frozen or fine-tuned backbone is passed through the existing classifier head to obtain logits \(z_{\text{cls}}=W_{\text{cls}} g\), which coincide with the baseline prediction (\texttt{cls\_logits}).

\textbf{Attribute-guided branch.} The attribute logits \(\ell\) are passed through a sigmoid to obtain scores \(p=\sigma(\ell)\in[0,1]^K\), which we interpret as attribute activations for the current image. These activations are then aggregated into the class space using the fixed matrix \(M\). For class \(c\), we compute a weighted average over its attribute set:
\begin{equation}
s_c=\frac{\sum_{k} M_{c,k}\, p_k\cdot p_k}{\sum_{k} M_{c,k}\, p_k+\epsilon},
\label{eq:class-score}
\end{equation}

where \(s=[s_1,\dots,s_C]\) are class scores that are subsequently scaled to logits, yielding the attribute-guided output \(z_{\text{attr}}\).

\textbf{Adaptive fusion (low-cost self-correction).} To combine the two decisions, we use a learnable fusion weight \(\alpha\) and temperature \(\tau\) that rescale and interpolate the two logits. The final fused logits are
\begin{equation}
z_{\text{fus}}=\alpha\cdot (z_{\text{cls}}/\tau) + (1-\alpha)\cdot (z_{\text{attr}}/\tau).
\label{eq:fusion}
\end{equation}

During training, we always use the fused output \(z_{\text{fus}}\) as the sole classification head, computing the cross-entropy loss only on \(z_{\text{fus}}\); the attribute prediction and regularization terms act as auxiliary supervision to shape the shared representation and fusion weights, as validated by the ablations in Table~\ref{tab:ablation}.

\subsection{Joint Optimization Objective and Training Strategy}

The training objective combines a cross-entropy loss on the fused logits, an attribute prediction loss, and a regularization term that aligns attribute predictions with the class--attribute matrix:
\begin{equation}
\mathcal{L}= \lambda_{\text{fus}}\,\mathcal{L}_{\text{fus}} + \lambda_{\text{reg}}\,\mathcal{L}_{\text{reg}} + \lambda_{\text{attr-pred}}\,\mathcal{L}_{\text{attr-pred}}.
\label{eq:overall-loss}
\end{equation}

where \(\mathcal{L}_{\text{fus}}\) is the cross-entropy loss between the fused logits \(z_{\text{fus}}\) and the true class label, \(\mathcal{L}_{\text{attr-pred}}\) is the BCEWithLogits loss on attribute logits \(\ell\) against the binary target vector \(m_y\), and \(\mathcal{L}_{\text{reg}}\) is the regularization term in Eq.~\eqref{eq:reg-loss} that combines an MSE penalty and a cosine-similarity term between the predicted attribute activations (after sigmoid) and the same target vector \(m_y\) derived from \(M\). Using both terms encourages the model to match each attribute probability numerically while also aligning the overall attribute direction in the semantic space. All modules---including the backbone, attribute branch and fusion weight---are optimized jointly using a single optimizer, while the primary supervision for classification is always applied to \(z_{\text{fus}}\).

\section{Experiments}

\subsection{Datasets and Setup}

\textbf{Datasets.} We consider both public and private ultrasound datasets. The breast cancer three-class task is based on BUSI\cite{ref2} (normal/benign/malignant), while the fetal plane 7-class task uses an internal multi-center dataset of standard views (abdomen, brain, femur, etc.).

\textbf{Models and Training.} We adopt ImageNet-pretrained ResNet50, ViT-B and Vim-s-16 (BU-Mamba\cite{ref29}) backbones and simply attach our attribute-guided branch and fusion head, denoting the resulting model as AttrGuide. All models are trained with Adam under the same hyper-parameters (learning rate \(1\times10^{-4}\), batch size 32), and the attribute tables are designed by ultrasound experts from public guidelines for BUSI and the fetal views.

\subsection{Results and Analysis}

We first evaluate AttrGuide as a plug-in to an existing SOTA framework. On BUSI, we attach our module to the BU-Mamba Vim-s-16 encoder\cite{ref29} and compare it with ViT and VMamba baselines to verify that it works across diverse backbones without re-designing the architecture. We further test whether the same module remains effective in a multi-task setting where BUSI classification is trained jointly with segmentation or auxiliary tasks~\cite{ref34}. Table~\ref{tab:busi_encoders} reports BUSI test accuracy, including BU-Mamba with and without AttrGuide, and the accompanying multi-task plot shows how attribute guidance influences ACC and Macro F1.

\begin{table}[t]
\caption{Comparison on BUSI and multi-task BUSI setting. Left: BUSI test accuracy (mean $\pm$ std over 5 runs) for baselines and BU-Mamba with/without AttrGuide; $^{\dagger}$ denotes results reported in BU-Mamba \cite{ref29}. Right: multi-task ACC/F1 on the joint segmentation--classification framework \cite{ref34}.}
\label{tab:busi_encoders}
\centering
\small
\setlength{\tabcolsep}{4pt}
\renewcommand{\arraystretch}{1.05}
\begin{minipage}{0.38\textwidth}
  \centering
  \begin{tabular}{lc}
  \toprule
  Encoder & BUSI ACC \\
  \midrule
  ResNet50$^{\dagger}$ & 85.64 $\pm$ 2.72 \\
  VGG16$^{\dagger}$ & 85.47 $\pm$ 4.59 \\
  ViT-ti-16$^{\dagger}$ & 85.98 $\pm$ 4.48 \\
  ViT-s-16$^{\dagger}$ & 86.50 $\pm$ 4.44 \\
  ViT-s-32$^{\dagger}$ & 84.10 $\pm$ 3.81 \\
  ViT-b-16$^{\dagger}$ & 87.18 $\pm$ 2.70 \\
  ViT-b-32$^{\dagger}$ & 85.98 $\pm$ 1.39 \\
  BU-Mamba$^{\dagger}$ & \underline{87.86 $\pm$ 2.72} \\
  \midrule
  AttrGuide (Ours) & \textbf{88.72 $\pm$ 1.90} \\
  \bottomrule
  \end{tabular}
  {\centering\small (a) BUSI test accuracy for competing methods.\par}
\end{minipage}\hfill
\begin{minipage}{0.56\textwidth}
  \centering
  \includegraphics[width=\linewidth]{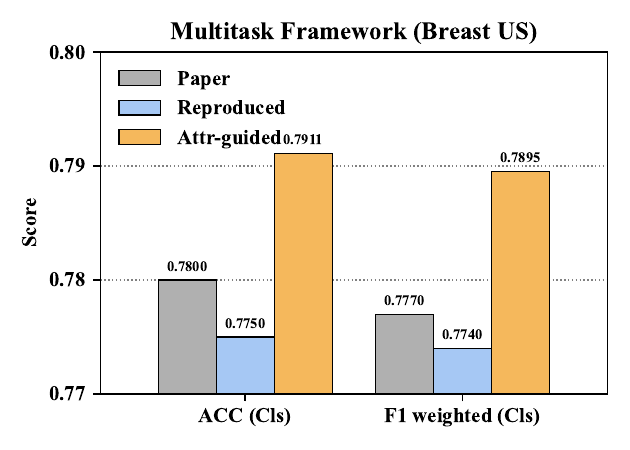}
  {\centering\small (b) Multi-task ACC and F1 \cite{ref34}.\par}
\end{minipage}
\end{table}

\begin{table}[t]
\caption{Overall performance on BUSI and fetal 7-class tasks. Acc (\%) and Macro F1 (\%) for Baseline and +Attribute Guidance ($\uparrow$ denotes absolute improvement over Baseline).}
\label{tab:macro_metrics}
\centering
\small
\setlength{\tabcolsep}{4pt}
\renewcommand{\arraystretch}{1.05}
\begin{tabular}{ll S[table-format=2.1] S[table-format=2.1] S[table-format=2.2] S[table-format=2.2]}
\toprule
Dataset & Backbone & {Acc} & {Acc} & {Macro F1} & {Macro F1} \\
 &  & {Baseline} & {+Attr} & {Baseline} & {+Attr} \\
\midrule
\multirow{2}{*}{BUSI (3-class)} & ViT-B & 81.1 & {\bfseries 85.4}\textcolor{green!50!black}{\scriptsize$\uparrow$4.3} & 79.86 & {\bfseries 82.76}\textcolor{green!50!black}{\scriptsize$\uparrow$2.90} \\
 & ResNet50 & 81.5 & {\bfseries 83.3}\textcolor{green!50!black}{\scriptsize$\uparrow$1.8} & 81.04 & {\bfseries 82.35}\textcolor{green!50!black}{\scriptsize$\uparrow$1.31} \\
\midrule
\multirow{2}{*}{Fetal (7-class, private)} & ViT-B & 92.8 & {\bfseries 94.9}\textcolor{green!50!black}{\scriptsize$\uparrow$2.1} & 90.69 & {\bfseries 94.25}\textcolor{green!50!black}{\scriptsize$\uparrow$3.56} \\
 & ResNet50 & 92.2 & {\bfseries 92.6}\textcolor{green!50!black}{\scriptsize$\uparrow$0.4} & 90.22 & {\bfseries 91.51}\textcolor{green!50!black}{\scriptsize$\uparrow$1.29} \\
\bottomrule
\end{tabular}
\end{table}

\begin{table}[t]
\caption{Training cost on BUSI. Time is minutes per fold and Params are in millions. For each backbone, $\Delta$ denotes the absolute increase of +Attr over Baseline ($\uparrow$).}
\label{tab:cost_busi}
\centering
\small
\setlength{\tabcolsep}{4pt}
\renewcommand{\arraystretch}{1.05}
\begin{tabular}{ll S[table-format=2.2] S[table-format=1.2] S[table-format=2.1] S[table-format=1.1]}
\toprule
Backbone & Setting & {Time} & {$\Delta$} & {Params} & {$\Delta$} \\
 &  & {(min)} & {(min)} & {(M)} & {(M)} \\
\midrule
\multirow{2}{*}{ViT-B}    & Baseline & 11.02 & \multicolumn{1}{c}{--} & 85.8 & \multicolumn{1}{c}{--} \\
                          & +Attr    & 11.18 & \textcolor{green!50!black}{\scriptsize$\uparrow$0.16} & 86.4 & \textcolor{green!50!black}{\scriptsize$\uparrow$0.6} \\
\midrule
\multirow{2}{*}{ResNet50} & Baseline & 10.35 & \multicolumn{1}{c}{--} & 23.5 & \multicolumn{1}{c}{--} \\
                          & +Attr    & 11.20 & \textcolor{green!50!black}{\scriptsize$\uparrow$0.85} & 24.8 & \textcolor{green!50!black}{\scriptsize$\uparrow$1.3} \\
\midrule
\multirow{2}{*}{Mamba}    & Baseline & 73.98 & \multicolumn{1}{c}{--} & 25.4 & \multicolumn{1}{c}{--} \\
                          & +Attr    & 75.91 & \textcolor{green!50!black}{\scriptsize$\uparrow$1.93} & 25.9 & \textcolor{green!50!black}{\scriptsize$\uparrow$0.5} \\
\bottomrule
\end{tabular}
\end{table}

\noindent\textbf{Generality.}
AttrGuide consistently improves performance across datasets and backbones, and remains effective when plugged into a strong baseline (BU-Mamba), improving BUSI test accuracy from 87.86\% to 88.72\% (Tables~\ref{tab:busi_encoders} and~\ref{tab:macro_metrics}).

\noindent\textbf{Low cost.}
The plug-in attribute-guided branch adds only marginal training-time and parameter overhead on BUSI (Table~\ref{tab:cost_busi}), while delivering consistent gains across backbones (Table~\ref{tab:macro_metrics}).

\begin{table}[t]
\centering
\caption{Ablation on the private fetal 7-class task (ViT-B). $\Delta$ denotes absolute improvement over Cls-only.}
\label{tab:ablation}
\small
\setlength{\tabcolsep}{6pt}
\renewcommand{\arraystretch}{1.05}
\begin{tabular}{ccc S[table-format=2.1] c}
\toprule
Cls branch & Attr branch & Learnable fusion & {Test Acc (\%)} & $\Delta$ \\
\midrule
\checkmark &             &                  & 89.1 & \multicolumn{1}{c}{--} \\
\checkmark & \checkmark  & Avg              & 90.3 & \textcolor{green!50!black}{\scriptsize$\uparrow$1.2} \\
           & \checkmark  &                  & 92.6 & \textcolor{green!50!black}{\scriptsize$\uparrow$3.5} \\
\checkmark & \checkmark  & \checkmark       & {\bfseries 94.9} & \textcolor{green!50!black}{\scriptsize$\uparrow$5.8} \\
\bottomrule
\end{tabular}
\end{table}

\begin{figure}[t]
\centering
\includegraphics[width=0.95\textwidth,trim=0 200 0 0,clip]{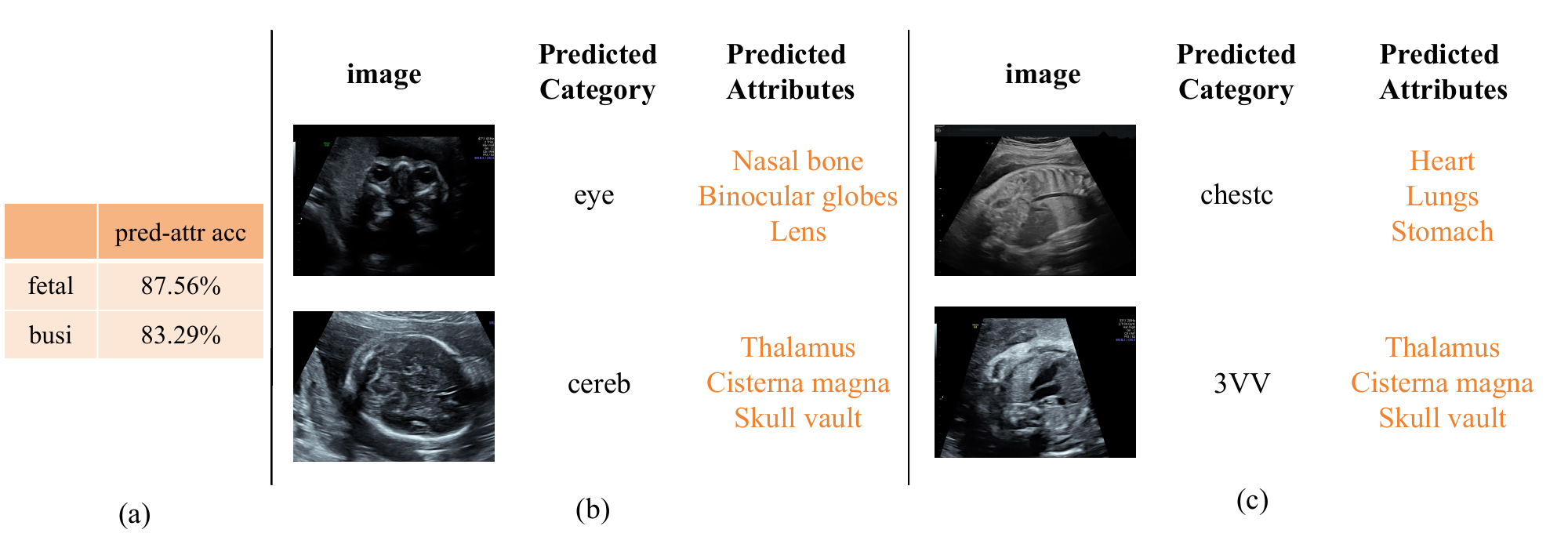}
\caption{Interpretability via attribute prediction. (a) Pred-attr accuracy: fetal 87.56\%, BUSI 83.29\%. (b)--(c) Example cases.}
\label{fig:vision}
\end{figure}

\noindent\textbf{Interpretability.}
Figure~\ref{fig:vision} shows that the attribute-guided branch predicts clinically meaningful attributes with high accuracy (87.56\% fetal, 83.29\% BUSI), providing human-interpretable cues for verification.

\noindent\textbf{Ablation.}
Table~\ref{tab:ablation} verifies that learnable fusion is crucial: the full model achieves the best accuracy (+5.8 over Cls-only) and outperforms naive averaging, showing that both branches are needed.

\FloatBarrier

\section{Conclusion}

In this paper, we introduce an attribute-guided dual-branch framework to enhance ultrasound classification by bridging visual features with clinical priors. We demonstrate that our approach consistently boosts the performance of SOTA backbones---for instance, improving BU-Mamba's accuracy on the BUSI dataset from 87.86\% to 88.72\%---while remaining highly efficient with less than 5\% additional training time. From the results, we mainly conclude that: (1) the injection of domain-agnostic attributes provides essential evidence that bolsters both model robustness and clinical interpretability; and (2) the adaptive fusion module is critical for synergistically integrating global and attribute-based semantics. In the future, we intend to extend this low-cost, plug-and-play enhancement to a broader range of medical imaging tasks.


\clearpage

\begin{thebibliography}{30}

\bibitem{ref1}
Burgos-Artizzu, X.P., Coronado-Gutiérrez, D., Valenzuela-Alcaraz, B., et al.:
Evaluation of deep convolutional neural networks for automatic classification of
common maternal fetal ultrasound planes. \textit{Scientific Reports} \textbf{10}, 10200 (2020).

\bibitem{ref2}
Al-Dhabyani, W., et al.:
Dataset of breast ultrasound images. \textit{Data in Brief} \textbf{28}, 104863 (2020).

\bibitem{ref3}
Chen, Y., Zhao, S., Chen, B., Gustaf, M.:
Clinically guided adaptive contrast adjustment for fetal plane classification:
a modular plug-and-play solution. \textit{Frontiers in Physiology} \textbf{16}, 1689936 (2025).

\bibitem{ref4}
Litjens, G., et al.:
A survey on deep learning in medical image analysis.
\textit{Medical Image Analysis} \textbf{42}, 60--88 (2017).

\bibitem{ref5}
Tajbakhsh, N., et al.:
Convolutional Neural Networks for Medical Image Analysis: Full Training or Fine Tuning?
\textit{IEEE Transactions on Medical Imaging} \textbf{35}(5), 1299--1312 (2016).

\bibitem{ref6}
Azizi, S., et al.:
Big self-supervised models advance medical image classification.
In: \textit{Proceedings of ICCV}, pp.~3478--3488 (2021).

\bibitem{ref7}
Shakeri, F., et al.:
Few-shot Adaptation of Medical Vision-Language Models.
In: \textit{Proceedings of MICCAI} (2024).

\bibitem{ref8}
Huang, Y., Cheng, P., Tam, R., Tang, X.:
Fine-grained Prompt Tuning: A Parameter and Memory Efficient Transfer Learning Method
for High-resolution Medical Image Classification.
In: \textit{Proceedings of MICCAI} (2024).

\bibitem{ref9}
Hussein, N., Shamshad, F., Naseer, M., Nandakumar, K.:
PromptSmooth: Certifying Robustness of Medical Vision-Language Models via Prompt Learning.
In: \textit{Proceedings of MICCAI} (2024).

\bibitem{ref10}
Zech, J.R., et al.:
Variable generalization performance of a deep learning model to detect pneumonia in chest radiographs:
a cross-sectional study. \textit{PLoS Medicine} \textbf{15}(11), e1002683 (2018).

\bibitem{ref11}
Chen, T., et al.:
A simple framework for contrastive learning of visual representations.
In: \textit{Proceedings of ICML}, pp.~1597--1607 (2020).

\bibitem{ref12}
He, K., et al.:
Momentum contrast for unsupervised visual representation learning.
In: \textit{Proceedings of CVPR}, pp.~9729--9738 (2020).

\bibitem{ref13}
You, K., Lee, S., Jo, K., Park, E., Kooi, T., Nam, H.:
Intra-class contrastive learning improves computer aided diagnosis of breast cancer in mammography.
In: \textit{Medical Image Computing and Computer Assisted Intervention -- MICCAI 2022},
pp.~331--340 (2022).

\bibitem{ref14}
Zheng, X., et al.:
XFMamba: Cross-Fusion Mamba for Multi-View Medical Image Classification.
arXiv:2503.02619 (2025).

\bibitem{ref15}
Feng, Z., Fu, J., Zou, X., Ye, H., Wu, H., Zhou, J., Wang, Y.:
Hybrid-View Attention Network for Clinically Significant Prostate Cancer Classification
in Transrectal Ultrasound. arXiv:2507.03421 (2025).

\bibitem{ref16}
Lampert, C.H., Nickisch, H., Harmeling, S.:
Attribute-based classification for zero-shot visual object categorization.
\textit{IEEE Trans. Pattern Anal. Mach. Intell.} \textbf{36}(3), 453--465 (2014).

\bibitem{ref17}
Lei, Y., Li, Z., Shen, Y., Zhang, J., Shan, H.:
CLIP-Lung: Textual knowledge-guided lung nodule malignancy prediction.
In: \textit{Medical Image Computing and Computer Assisted Intervention -- MICCAI 2023},
pp.~403--412 (2023).

\bibitem{ref18}
Ghosh, S., Poynton, C.B., Visweswaran, S., Batmanghelich, K.:
Mammo-CLIP: A Vision Language Foundation Model to Enhance Data Efficiency and Robustness in Mammography.
In: \textit{Proceedings of MICCAI} (2024); arXiv:2405.12255.

\bibitem{ref19}
Radford, A., et al.:
Learning transferable visual models from natural language supervision.
arXiv:2103.00020 (2021).

\bibitem{ref20}
Gao, Y., Gu, D., Zhou, M., Metaxas, D.:
Aligning Human Knowledge with Visual Concepts Towards Explainable Medical Image Classification.
arXiv:2406.05596 (2024).

\bibitem{ref21}
Fang, X., Lin, Y., Zhang, D., Cheng, K.-T., Chen, H.:
Aligning Medical Images with General Knowledge from Large Language Models.
arXiv:2409.00341 (2024).

\bibitem{ref22}
Koh, P.W., Nguyen, T., Tang, Y.S., Mussmann, S., Pierson, E., Kim, B., Liang, P.:
Concept Bottleneck Models. arXiv:2007.04612 (2020).

\bibitem{ref23}
Oikarinen, T., Das, S., Nguyen, L.M., Weng, T.-W.:
Label-Free Concept Bottleneck Models. arXiv:2304.06129 (2023).

\bibitem{ref24}
Yuksekgonul, M., Wang, M., Zou, J.:
Post-hoc Concept Bottleneck Models. arXiv:2205.15480 (2023).

\bibitem{ref26}
Kittler, J., Hatef, M., Duin, R.P.W., Matas, J.:
On combining classifiers. \textit{IEEE Trans. Pattern Anal. Mach. Intell.} \textbf{20}(3), 226--239 (1998).

\bibitem{ref27}
Guo, S., Wang, L., Chen, Q., Wang, L., Zhang, J., Zhu, Y.:
Multimodal MRI image decision fusion-based network for glioma classification.
\textit{Frontiers in Oncology} \textbf{12}, 819673 (2022).

\bibitem{ref29}
Nasiri-Sarvi, A., Hosseini, M.S., Rivaz, H.:
Vision Mamba for Classification of Breast Ultrasound Images.
arXiv:2407.03552 (2024). (MICCAI 2024 Deep-Breath Workshop).

\bibitem{ref30}
Lin, Z., et al.:
UniUSNet: A Promptable Framework for Universal Ultrasound Disease Prediction and Tissue Segmentation.
arXiv:2406.01154 (2024).

\bibitem{ref32}
Chen, S., Wang, W., Xia, B., et al.:
TransZero: Attribute-Guided Transformer for Zero-Shot Learning.
arXiv:2112.01683 (2021).

\bibitem{ref34}
Aumente-Maestro, C., Díez, J., Remeseiro, B.:
A multi-task framework for breast cancer segmentation and classification in ultrasound imaging.
\textit{Computer Methods and Programs in Biomedicine} \textbf{260}, 108540 (2025).

\end{thebibliography}
\end{document}